%% file: main.tex
\documentclass[10pt,twocolumn,letterpaper]{article}

\usepackage[pagenumbers]{cvpr} % To force page numbers, e.g. for an arXiv version

\input{preamble}

\definecolor{cvprblue}{rgb}{0.21,0.49,0.74}
\usepackage[pagebackref,breaklinks,colorlinks,citecolor=cvprblue]{hyperref}
\def\paperID{13082} % *** Enter the Paper ID here
\def\confName{CVPR}
\def\confYear{2024}

\definecolor{Gray}{gray}{0.9}
\title{SoundingActions: Learning How Actions Sound from Narrated Egocentric Videos}

\author{
Changan Chen$^{1}$ \hspace{2mm} Kumar Ashutosh$^{1,2}$ \hspace{2mm} Rohit Girdhar$^{2}$ \hspace{2mm} David Harwath$^{1}$ \hspace{2mm} Kristen Grauman$^{1,2}$  \\
$^1$University of Texas at Austin \hspace{3mm} $^2$FAIR, Meta
}

\begin{document}
\maketitle

\input{sections/abstract}

\input{sections/intro}
\input{sections/related_work}
\input{sections/task}
\input{sections/approach}
\input{sections/evaluation}
\input{sections/experiments}
\input{sections/conclusion}

{
    \small
    \bibliographystyle{ieeenat_fullname}
    \bibliography{main}
}
\clearpage
\input{sections/supp}


\end{document}

%% file: preamble.tex
\usepackage[dvipsnames]{xcolor}

\usepackage[utf8]{inputenc} % allow utf-8 input
\usepackage[T1]{fontenc}    % use 8-bit T1 fonts
\usepackage{url}            % simple URL typesetting
\usepackage{booktabs}       % professional-quality tables
\usepackage{amsfonts}       % blackboard math symbols
\usepackage{nicefrac}       % compact symbols for 1/2, etc.
\usepackage{microtype}      % microtypography

\usepackage{graphicx}
\usepackage{amssymb}% http://ctan.org/pkg/amssymb
\usepackage{pifont}% http://ctan.org/pkg/pifont
\usepackage{multirow}
\usepackage{amsmath}
\usepackage{color, colortbl}
\usepackage{caption}
\usepackage{subcaption}
\usepackage{siunitx}
\definecolor{citecolor}{HTML}{0071bc}

\definecolor{Gray}{gray}{0.9}
\newcommand{\KGnote}[1]{{\color{red}{\bf KG: }#1}}

\newcommand\KGCR[1]{\textcolor{blue}{#1}}
\newcommand\cc[1]{\textcolor{magenta}{#1}}

\newcommand\CAcr[1]{\textcolor{teal}{#1}}
\newcommand\CAr[1]{\textcolor{black}{#1}} %teal
\newcommand{\CAnote}[1]{{\color{brown}{\bf CA: }#1}}

\newcommand{\cmark}{\ding{51}}%
\newcommand{\xmark}{\ding{55}}%
\newcommand{\tb}{\bfseries}
\newcommand{\RNum}[1]{\uppercase\expandafter{\romannumeral #1\relax}}

\renewcommand{\CAnote}[1]{}
\renewcommand{\KGnote}[1]{}

\renewcommand\cc[1]{\textcolor{magenta}{#1}}

\renewcommand\cc[1]{\textcolor{black}{#1}}

\renewcommand\KGCR[1]{\textcolor{black}{#1}}
\renewcommand\CAcr[1]{\textcolor{black}{#1}}

%% file: sections/abstract.tex
\begin{abstract}

We propose a novel self-supervised embedding to learn \emph{how actions sound} from narrated in-the-wild egocentric videos.  Whereas existing methods rely on curated data with known audio-visual correspondence, our multimodal contrastive-consensus coding (MC3) embedding reinforces the associations between audio, language, and vision when all modality pairs agree, while diminishing those associations when any one pair does not. We show our approach can successfully discover how 
\KGCR{the long tail of} human actions sound
\KGCR{from} egocentric video, outperforming an array of recent multimodal embedding techniques on two datasets (Ego4D and EPIC-Sounds) and multiple cross-modal tasks.  
\end{abstract}

%% file: sections/intro.tex
\section{Introduction}\label{sec:intro}

Human activity often produces sounds.  Closing a door, chopping vegetables, typing on a keyboard, talking with a friend---our interactions with the objects and people around us generate audio that reveals our physical behaviors.  These sounds can be strongly associated with the subjects of our activity and how we perform it.  For example, opening a water bottle sounds different than opening a cabinet; chopping sweet potatoes sounds different than chopping onions; chopping onions sounds different than mincing onions (the same object).  Understanding the link between sounds and actions is valuable for a number of applications, such as multimodal activity recognition, cross-modal retrieval, content generation, or forecasting the physical effects of a person's actions.

How should AI learn about \emph{sounding actions}? Existing work typically curates annotated datasets for supervised learning~\cite{audioset,EPICSOUNDS2023,vgg_sound,piczak2015dataset}, taking care to select events or actions that have associated sounds (e.g., lawnmowing, chopping), while others deliberately collect videos of object collisions (e.g., striking objects with a drumstick~\cite{owens2016vis} or crashing into them with a robot~\cite{Gandhi2020swoosh,clarke2021diffimpact}), or develop physics-based simulations~\cite{tdw}. 
On the one hand, these approaches are appealing for their ability to focus on meaningful audio-visual correspondences.  On the other hand, their curated nature risks limiting the scope of sounding actions that can be learned.

\begin{figure}[t]
    \centering
    \includegraphics[width=\linewidth]{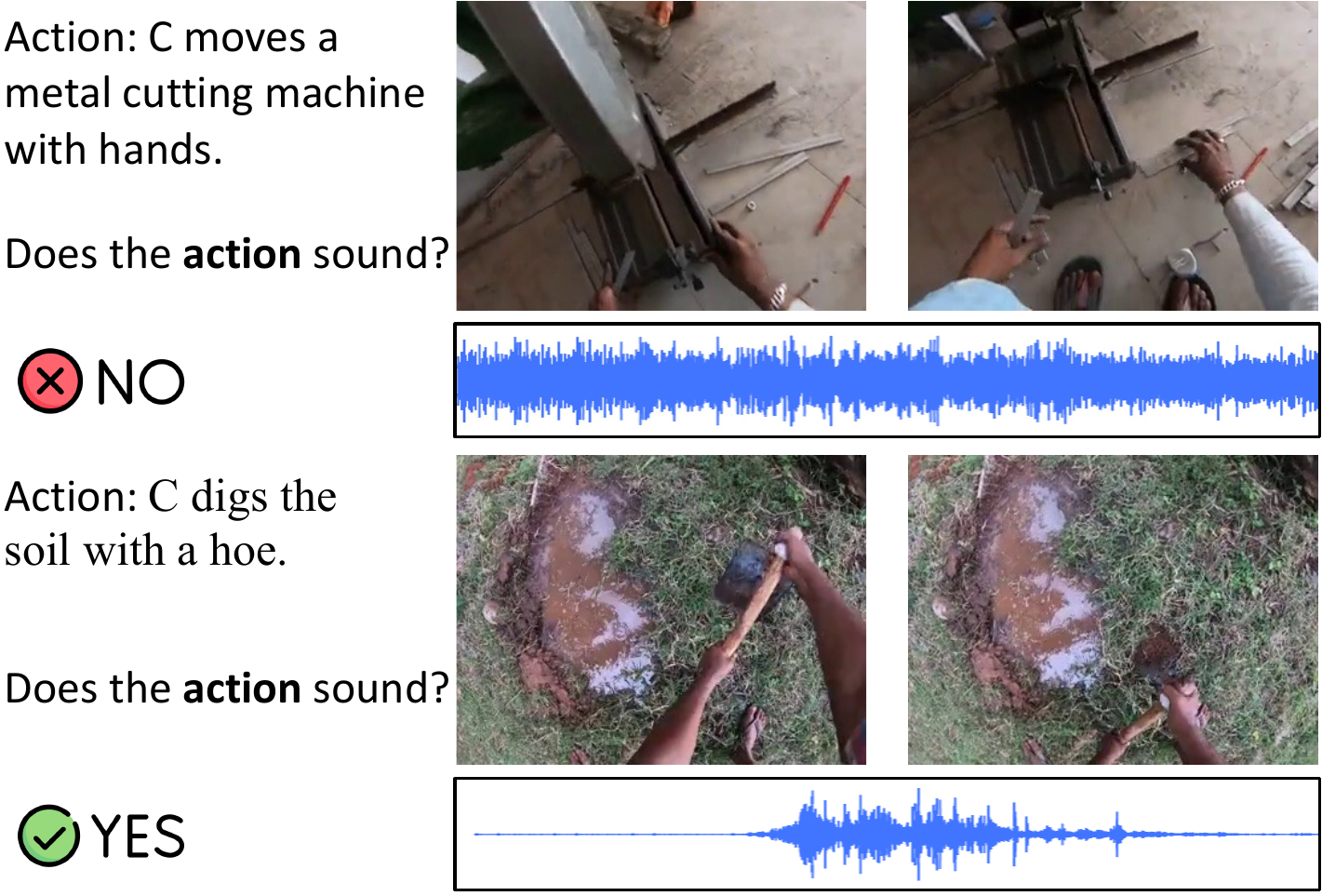}
    \caption{We aim to distinguish sounds that are directly caused by human actions (bottom) from those that are not (top).  Given egocentric training videos with language descriptions of the camera wearer's (``C") current action, we learn an embedding where the audio and visual features of any given clip are best aligned only when both are also consistent with the language.  This allows discerning clips where the audio and vision may be \emph{correlated} (e.g., the cutting machine running making loud noise in top row) versus those where the sounds are \emph{driven by human action} (digging in bottom row)---importantly, without language at inference time. 
    }
    \label{fig:pull_figure}
    % \vspace{-0.1in}
\end{figure}

\begin{figure*}[t]
    \centering
    \includegraphics[width=\linewidth]{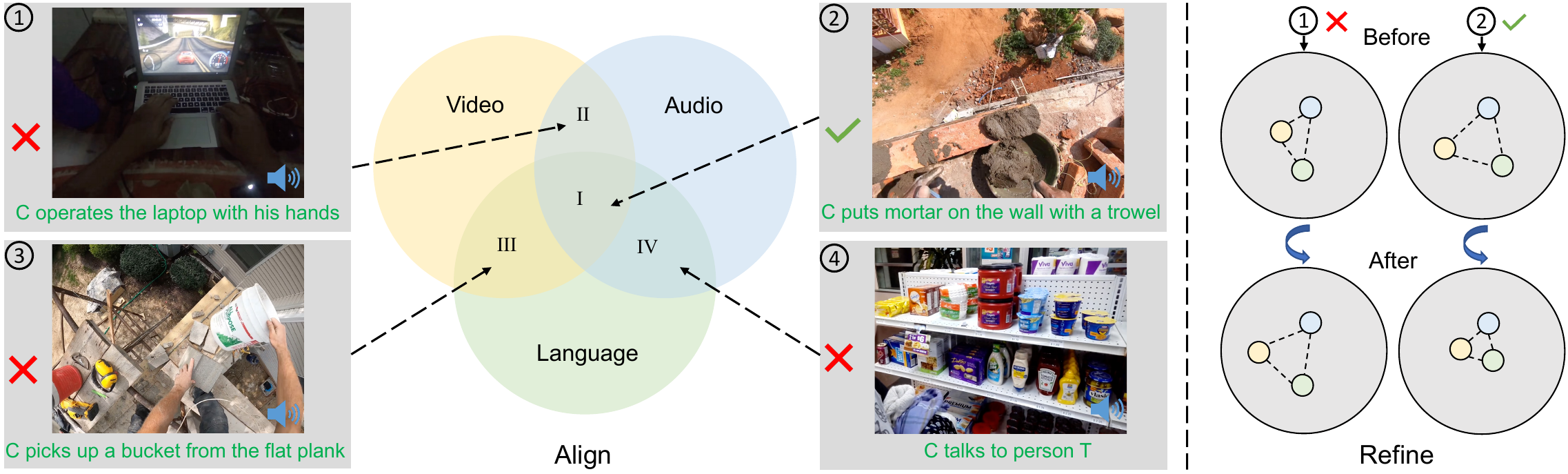}
    \caption{\small{\textbf{Main idea}.
On the left, the Venn diagram illustrates different ways audio ($A$), video ($V$) and language ($L$) modalities can overlap in the content they capture. C refers to the camera wearer. Regions \RNum{2},\RNum{3},\RNum{4} are information that is only shared between two modalities but not the third, e.g., the racing game in \textcircled{\raisebox{-0.9pt}{1}} where the game} sounds correlate with the vision, yet are not about the camera wearer's described action (using hands on laptop), the lifting action in \textcircled{\raisebox{-0.9pt}{3}}, where the visuals and language agree but the action is inaudible, and the \KGCR{off-screen} %invisible
talking action in \textcircled{\raisebox{-0.9pt}{4}}, where talking is heard and described, but the camera wearer cannot be seen speaking. Region \RNum{1} is the information that corresponds to all modalities agreeing, e.g., the visible and audible plastering action in  \textcircled{\raisebox{-0.9pt}{2}}.  
    Our model's ``align" phase detects any such (dis)agreements via pairwise contrastive learning on the modalities. 
    In the ``refine" phase, we use the intersection of that agreement (region \RNum{1}) to refine the embedding.   For example, on the right, we show what the three modality embeddings should look like after the ``align" stage for examples 1 and 2. Embeddings of instances where all modalities agree will be closer in the embedding space and apart otherwise.  In other words, for example 1, yellow (video) cannot be close to blue (audio) unless green is too (language).
    }\label{fig:concept}
    % \vspace{-0.1in}
\end{figure*}

Instead, we aim to learn how human actions sound from narrated in-the-wild egocentric videos. See Figure~\ref{fig:pull_figure}.
Given a pool of videos of everyday human activity, 
the goal is to learn a cross-modal representation where sounding actions cluster together based on how they look and sound.  By sampling the videos freely, we can broaden the scope to discover %a long tail of non-obvious, 
the breadth of %everyday
sounding actions without having to rely on a closed, pre-defined set of action categories.  In particular, by focusing on \emph{unscripted egocentric} video from wearable cameras in daily-life settings~\cite{ego4d,Damen2018EPICKITCHENS}, we aim to include subtle and long-tail scenarios unavailable in curated datasets, such as sounds of keys jangling when unlocking a door, scissors snipping when cutting the dog's fur, or fingernails scratching on one's own arm. Egocentric video is a particularly attractive source here because 1) human interaction sounds are more audible in near-field egocentric recordings and 2) passively captured long-form ego-video simply covers more everyday sounds, including the rare ones.

However, the learning task is challenging because some visible actions do not make any sound, and some sounds are the result of off-screen actions. Finally, other sounds may be \emph{correlated} with on-screen objects (such as traffic noise and a city street), but are not directly related to the video by a salient and visible camera wearer action. For this reason, although existing self-supervised audio-visual methods~\cite{Arandjelovi2018look,Objects_sound,korbar-nips2018,owens2018audio,cm_acc,visual-echoes,imagebind,lorenzo-nips2020} are good at detecting audio-visual correspondences, they tend to capture general %correspondences
correlations
rather than the action-specific correspondence.

To address this challenge, we propose a novel \emph{multimodal consensus embedding} approach. 
Importantly, we suppose the in-the-wild egocentric training videos are accompanied by free-form natural language descriptions describing the actions of the camera wearer, as provided in the ``narrations" of existing large-scale ego-video datasets~\cite{ego4d,Damen2018EPICKITCHENS}.  
The main idea is to seek video samples where there is semantic agreement between all three modalities---the audio, visual, and language---while distancing those that do not.  This \emph{intersection} of the modalities with language assures that correspondences in the audio and visual streams stem from alignment on the sounding action.  

To achieve this, the proposed model first aligns a preliminary embedding from contrastive losses imposed per instance on each pair of modalities.  Next, we refine those embeddings with a consensus objective that targets a minimum (bottleneck) pairwise similarity.  The latter pushes all pairs of inter-modality agreement towards this consensus---or lack thereof---while jointly continuing to optimize the paired-modalities' contrastive losses.  In this way, we overcome the simplifying assumption made by existing multimodal embeddings that require all modalities to agree~\cite{cmc,imagebind,VATT}. See Figure~\ref{fig:concept}.

We demonstrate our approach by training with in-the-wild data from Ego4D~\cite{ego4d} without audio labels and testing on both Ego4D and EPIC-Sounds~\cite{EPICSOUNDS2023}. To allow a formal large-scale evaluation of sounding actions, we introduce a dataset of %71K
professional annotations on 33K video clips spanning Ego4D.
Our model successfully discovers sounding actions that agree with ground truth labels on both datasets.
Compared to existing multimodal embedding paradigms~\cite{clap,cm_acc,cmc,imagebind}, our model not only better discovers sounding actions and learns embeddings for cross-modality retrieval, but also generalizes \KGCR{better} to the audio classification benchmark on EPIC-Sounds.
To our knowledge, this is the first result of its kind to show sounding actions discovered organically from narrated in-the-wild video.  
% We will release the data and code for our models. \CAnote{legal requires to remove this sentence}

%%%\KGnoteCR{will these be up by next week?} \CAnotecr{yes, I can get that ready by next week}

%% file: sections/related_work.tex
% \vspace{-0.1in}
\section{Related Work} \label{sec:related_work}

\paragraph{Action/interaction/impact sound.} 
Some work~\cite{kazakos2019TBN,Nagrani21c,gao2020listentolook} leverages audio to improve activity recognition on video datasets such as UCF101~\cite{ucf} and ActivityNet~\cite{caba2015activitynet}, which have visual labels but no audio labels. Existing audio datasets such as AudioSet~\cite{audioset} and VGG-Sound~\cite{vgg_sound} 
target general sound classes such as music, speech, and sports. 
EPIC-Sounds~\cite{EPICSOUNDS2023} provides an audio classification benchmark for actions in kitchen environments, but it has no labels for the correspondence between the visual action and the sound.
The Greatest Hits dataset~\cite{owens2016vis} contains videos where people hit and scratch object surfaces with a drumstick, which enables audio synthesis from videos. 
Interaction sound has also been studied in robotics, e.g., using a robotic platform to collect sounds and study the synergy between action and sounds~\cite{Gandhi2020swoosh,clarke2021diffimpact}. Impact sounds are modeled in a physics-based simulator~\cite{tdw}. 
\KGCR{Throughout, the} existing work assumes a fixed, given taxonomy of action classes or audio labels of interest.
In contrast, we learn how actions make sounds from in-the-wild narrated egocentric videos, 
without relying on a taxonomy of discrete labels for the audio events.

\vspace{-0.15in}
\paragraph{Audio-visual learning.} 
As naturally co-occurring data, there are rich correspondences between video and audio, such as spatial correspondence~\cite{25d-visual-sound,morgado-spatial-nips2020,hao2022asl}, acoustic correspondence~\cite{singh_image2reverb_2021,vam2022}, semantic correspondence~\cite{Objects_sound,Arandjelovi2018look}, and lip motion~\cite{gao_visualvoice_2021,shi2022avhubert}. Existing work typically either learns one type of correspondences by manually creating misalignment, e.g., shifting audio temporally to create temporal supervision~\cite{owens2018audio,korbar-nips2018,Arandjelovi2018look} or
down-mixing
audio channels to create spatial supervision~\cite{morgado-spatial-nips2020,25d-visual-sound}, or it picks up any correspondence that emerges from the learning process~\cite{morgado2021_robust_xid,mittal2022learning} (e.g., actions, objects, environments). 
In contrast, 
we aim to learn action-specific correspondence by leveraging the semantic grounding from human narrations.

\vspace{-0.15in}
\paragraph{Language+X learning.}
Language can provide semantic grounding for what we see or hear. Prior work exploring language with another modality includes 
image/video captioning~\cite{show_and_tell}, visual question answering~\cite{VQA}, and audio captioning
~\cite{Mei2021act}. Recent results with large-scale image-text datasets %has been shown 
show that language is excellent for guiding the learning of image features (such as CLIP~\cite{clip}), video features~\cite{kevin2022egovlp,zhao2023lavila}, or audio features~\cite{guzhov2021audioclip}. However, there is limited work studying binding language to more than one modality, as we propose. 
Some recent work~\cite{zellers2022merlotreserve,VATT} explores self-supervised learning with language, vision, and sound, where the language is typically the transcription of the audio. They construct self-supervised objectives under the assumption that modalities agree with each other. In contrast, we use language that is different from the speech transcription and we %also
explore modality agreement in training.

\vspace{-0.15in}
\paragraph{Multi-modal/view representation learning.} Recent work builds multimodal models using more than two modalities (e.g., audio, video, and language) for improving the representations of one modality~\cite{shvetsova2022everything,alayrac2020self,VATT}, while others explore modality-invariant or view-invariant features
~\cite{tcgm,cmc,imagebind}. 
 For example, ImageBind~\cite{imagebind} binds the visual features with other modalities in sequence, while 
CMC~\cite{cmc} proposes a contrastive multi-view loss that maximizes the mutual information between different views of the same scene. These methods 
assume there exists shared information among all modalities, which does not address how to find these examples. Our method implicitly discovers the shared information by  \KGCR{analyzing} %leveraging 
the multimodal consensus, or lack thereof.

%% file: sections/task.tex
\vspace{-0.05in}
\section{Task Formulation}\label{sec:task}
%Learning How Actions Sound}}\label{sec:task}
\vspace{-0.05in}

We define a \KGCR{\textbf{sounding action}} as a human-initiated action that produces sound during its %the 
execution due to interactions with the surrounding environment. We are particularly interested in learning % how actions sound, especially 
how subtle and long-tail daily human actions sound. 
If hypothetically we were
given a clip with audio $a$, video $v$, and label $y$ indicating whether the clip contains a sounding action, our objective would be to minimize the distance between audio-visual embeddings if $y=1$ and maximize the distance between them if $y=0$, i.e.,  minimizing $(-1)^y \mathcal{D}(e_a, e_v)$, where $\mathcal{D}$ measures the distance and $e_{a,v}$ are their embeddings.
 However, we do not assume access to any such direct supervision; labeling sounding actions is expensive, both because many actions do not produce sounds, and because many clips do not contain actions. Instead, 
we aim to discover sounding actions in a weakly supervised %unsupervised 
fashion, while simultaneously learning 
  multimodal embeddings that capture them well.

To this end, we leverage ``narrations", a form of language description that is collected in recent egocentric video datasets such as Ego4D~\cite{ego4d} and EPIC-Kitchens~\cite{Damen2018EPICKITCHENS}.
These narrations %have two main characteristics: 1) they are timestamped, and 2) they are
are timestamped free-form sentences describing the current %foreground 
action being performed by the camera-wearer. 
 See
Figure~\ref{fig:concept} for examples.
Note that there may be other events in the video, too (e.g., a TV is playing), but these are \emph{not} narrated.  This is significant: the language specifically addresses near-field human interactions with objects, people, and the environment. 
The narrations offer two key benefits: % of using narration data: 
 1) the timestamps provide \emph{temporal} grounding of actions that occur in the video, indicating where potentially interesting clips are % so that it reduces the effort for curating clips, 
 and 2) the language provides \emph{semantic} grounding of actions---which our multimodal consensus idea will exploit %to train models 
 to learn action-specific audio-visual correspondence.

Formally, given a video with frames $v\in \mathcal{R}^{T\times H\times W\times C}$, audio $a\in \mathcal{R}^{S}$, and language narration $l$, where $T$ and $S$ are the number of frames for video and audio respectively, the goal is to learn embeddings $e_v$ and $e_a$ that are close in the embedding space if both $a$ and $v$ capture the same human action described in $l$, and distant otherwise.
If we plot how the three modalities overlap in a Venn diagram (Figure~\ref{fig:concept}), we can see that what we are interested in learning is exactly region \RNum{1}, i.e., a camera-wearer %foreground 
action that sounds.  From an information-theory perspective, this is equivalent to learning modality-invariant embeddings.

%% file: sections/approach.tex
% \vspace{-0.1in}
\section{Multimodal Contrastive-Consensus Coding} \label{sec:approach}
\begin{figure}[t]
    \centering
    \includegraphics[width=\linewidth]{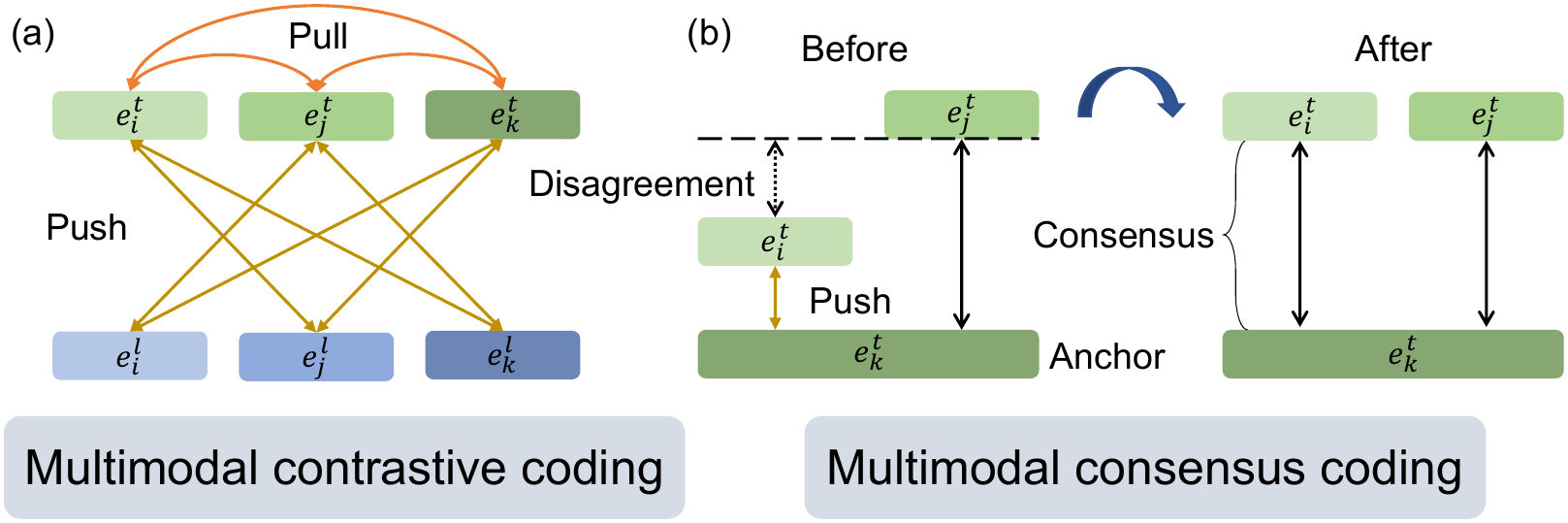}
    \caption{\textbf{Multimodal contrastive-consensus loss}. (a): Given three modality embeddings $e_i^t$, $e_j^t$, $e_k^t$, multimodal contrastive coding pulls each pair of modalities closer while pushing modality pairs from another sample further away. (b): However, not all modalities agree on how close they should be depending on the instance. Thus we set the furthest distance a feature has with respect to the anchor feature as the consensus and push the remaining embeddings away to meet this consensus. 
    % \rohit{I think you should put superscripts to show that all green are sample 1, blue is sample 2 etc}
    }
    \label{fig:loss}
% \vspace{-0.1in}
\end{figure}

% \vspace{-0.05in}
%In this section, 
Next we present our solution MC3 (Multimodal Contrastive-Consensus Coding) for learning modality-invariant embeddings, which consists of an \emph{inter-sample contrastive loss} and an \emph{intra-sample consensus loss}. See Fig.~\ref{fig:loss}. We first present the two-stage training framework in Sec.~\ref{sec:two_stage} and then discuss the two losses in Sec.~\ref{sec:contrastive} and Sec.~\ref{sec:consent}. For simplicity, we denote the $n$ input modalities as $M_i, i\in [1, n]$.

% \vspace{-0.05in}
\subsection{{Align-Refine} Two-stage Training}\label{sec:two_stage}
We design a two-stage training paradigm. The high-level idea is to first optimize the pairwise agreement in an ``align" stage, and then refine these embeddings with global consensus in the ``refine" stage. See Fig.~\ref{fig:concept}.

{In the first stage}, we train modality encoders with a contrastive loss $\mathcal{L}_{\text{contrastive}}$, which guides modality embeddings to have a good initial alignment that captures the pairwise similarity between modalities \KGCR{that capture the same underlying action}, as opposed to random initialization. 

In the second stage, we {refine the pairwise-aligned embeddings with a globally established consensus.}
\KGCR{Specifically,} we train the model with a consensus loss $\mathcal{L}_{\text{consensus}}$ that pushes all \CAcr{intra-sample} modality agreement towards this consensus, while jointly optimizing the contrastive loss $\mathcal{L}_{\text{contrastive}}$, to maximally capture the shared information across modalities. 
The MC3 loss $\mathcal{L_{\text{MC3}}}$ combines the contrastive and consensus losses, and will be detailed below. 
We confirm %show
experimentally that it is important to keep the contrastive loss in the second stage, although the main purpose of this stage is to refine embeddings with consensus.

\subsection{Multimodal Contrastive Coding}\label{sec:contrastive}
% \vspace{-0.05in}
{Cross-modal} contrastive learning has been shown to discover {representations} where modalities are informative of each other~\cite{morgado2021_robust_xid}. 
Prior work~\cite{cpc,variational_bound} shows that minimizing the contrastive loss between $M_i$ and $M_j$ maximizes the lower bound on the mutual information $I(M_i; M_j)$.
Inspired by this, we first use contrastive learning to optimize the pairwise similarities $\mathcal{S}(e_i, e_j)=e_i e_j$, where {$e_{i,j}$} is the latent embedding normalized on the unit sphere for modality pair {$i,j$}. 
We use the InfoNCE~\cite{oord2018representation} loss to optimize each individual $\mathcal{S}(e_i, e_j)$ as follows:
\begin{equation}\label{eq:contrastive}
    \mathcal{L}_{i,j} = {-}\frac{1}{|\mathcal{B}|} \sum_{t \in \mathcal{B}}\log \frac{\exp(e_i^t  e_j^t/\tau)}{\sum_{l \in \mathcal{B}}\exp(e_i^t e_j^l/\tau)},
\end{equation}
where $\mathcal{B}$ is the batch and $\tau$ is the temperature. This loss treats modalities from the same sample as positive pairs and pulls them closer and it treats modalities from different samples as negative pairs and pushes them apart. See Fig.~\ref{fig:loss} (a).  
The total loss is the sum of losses enumerated over all pairs of modalities, i.e., $\mathcal{L}_{\text{contrastive}} = \sum_{i,j}\mathcal{L}_{i,j}$.

% \vspace{-0.05in}
\subsection{Multimodal Consensus Coding}\label{sec:consent}
% \vspace{-0.05in}
The contrastive loss above attempts to bring all {temporally co-occurring} modalities closer assuming there are strong correspondences among them in the input space.  However, naively doing so would be problematic for instances where not all modalities agree (cf.~Figure~\ref{fig:concept}). To tackle this issue, we propose a novel objective that leverages the consensus of inter-sample modalities discovered from the contrastive coding as additional supervision.

First of all, we choose an anchor modality $M_a$, which serves as the point of comparison for other modalities $M_i, i \in [1, n], i \neq a$. With the normalized embedding $e_i^t$ of modality $i$ and sample $t$, we then compute the cosine similarity score between each non-anchor modality and the anchor modality.
Now, these similarity scores may or may not agree with each other. To only learn embeddings shared across all modalities, we set the consensus score as the minimum (bottleneck) score:
\begin{equation}
    c^t = \mathcal{K}^{-1}(\min_{i,i\neq a}(\mathcal{K}_1(e_1^t e_a^t),  ..., \mathcal{K}_n(e_n^t e_a^t))),
\end{equation}
where $\mathcal{K}_i(x) = ((x+1)/2)^{\alpha_i}, x\in [-1, 1]$ is a modality-specific scaling function {that first maps scores to [0, 1] and then adjusts the distribution with a tunable parameter $\alpha_i$. $\mathcal{K}^{-1}$ is the inverse function that maps the scaled score back to the original space. The intuition behind $\mathcal{K}_i(x)$ is that different modalities carry different amounts of information and we want} to normalize the score distributions among the different modality pairs, making them comparable.

The consensus score $c^t$ is high if and only if all pairwise scores are high, and it is low if there exists at least one modality that does not agree with the anchor modality. After obtaining the consensus score, we design a loss that forces all modalities to follow this consensus, as follows:
\begin{equation}\label{eq:consensus}
    \mathcal{L}_{\text{consensus}} = \frac{1}{|\mathcal{B}|} \sum_{t \in \mathcal{B}}\sum_{i, i\neq a} ||e_i^t e_a^t - c^t||_2
\end{equation}

The total loss $\mathcal{L_{\text{MC3}}}$ is the sum of the contrastive loss (Eq.~\ref{eq:contrastive}) and the consensus loss (Eq.~\ref{eq:consensus}):
\begin{equation}
\begin{aligned}
    \mathcal{L_{\text{MC3}}} 
    = {-}\frac{1}{|\mathcal{B}|} (&\sum_{t \in \mathcal{B}} \underbrace{\sum_{i,j}\log \frac{\exp(e_i^t e_j^t/\tau)}{\sum_{l \in \mathcal{B}}\exp(e_i^t e_j^l/\tau)}}_{\text{Inter-sample}} \\
    &\phantom{=} - \underbrace{\sum_{i, i\neq a} ||e_i^t e_a^t - c^t||_2}_{\text{Intra-sample}}).
\end{aligned}
\end{equation}
\KGCR{This loss} pushes embeddings with a low consensus score apart while pulling together embeddings with a high consensus score, and thus aligns embeddings better in the joint embedding space. See Fig.~\ref{fig:loss}. 

{Optimizing this loss is not trivial since it has both contrastive and reconstruction objectives. Indeed,} directly optimizing the loss does not work well as shown in the ablation study (\cref{sec:sounding_action_discovery}). %We devise the 
The proposed two-stage training paradigm (\cref{sec:two_stage}) helps train the model stably.

\subsection{Implementation Details}\label{sec:implementation}
Our modalities of interest are $M_1=A$ (audio), $M_2=V$ (vision), and $M_3=L$ (language). There are six pairwise contrastive losses for three modalities. 
When computing the modality consensus, we empirically find using audio as the anchor leads to the best results in our task (cf.~Sec.~\ref{sec:experiments}). {We set the scaling parameters $\alpha_l$ and $\alpha_v$} to $1$ and $0.5$ respectively, based on a hyperparameter search on the validation set. See ablations in Supp.

For extracting the feature representations, we use TimeSformer~\cite{gberta_2021_ICML} as our video encoder, DistillBERT~\cite{Sanh2019DistilBERTAD} as our text encoder, and AST~\cite{gong21b_interspeech} as our audio encoder. We initialize the video and language encoders with embeddings from ~\cite{kevin2022egovlp}, {and the audio encoder with embeddings pretrained on ImageNet~\cite{deng2009imagenet}}. We train all encoders. We choose these initial encoders due to their good results in the literature; however, our MC3 loss is not specific to the choice of these encoders and others could be swapped in.

We train all models on 8 A40 GPUs with a learning rate of $3\mathrm{e}{-5}$ and batch size of $256$ for $5$ epochs \CAcr{for both stages}, and take the final checkpoint for evaluation. We use the Adam optimizer~\cite{kingma2014adam}. Our implementation is based on the codebase from \cite{kevin2022egovlp}. 

% \KGnoteCR{please double check that all the items from Q2 to the first reviewer in rebuttal are accounted for.}

%% file: sections/evaluation.tex
% \vspace{-0.05in}
\section{Training and Eval Data for 
Sounding Actions} \label{sec:data}
% \vspace{-0.05in}

\paragraph{Dataset.} Ego4D~\cite{ego4d} is a large-scale egocentric video dataset that has more than 3,600 hours of video recordings depicting hundreds of daily activities---and 2,113 of those hours have audio available.  As discussed, it also has time-stamped narrations \cc{that are free-form sentences describing the current activity performed by the camera-wearer}. 
However, 
Ego4D has no annotation of whether an action makes sounds, what sounds an action makes, 
or whether there exists other (non-action) sounds.
It is thus non-trivial to detect if an action in the clip makes sound based on simple heuristics, e.g., the burst of sound energy, since many actions %are complicated and 
could produce continuous sounds %that present
with ambient-sound characteristics, e.g., wiping tables or sawing wood.

We construct the training dataset by 
extracting clips from each Ego4D video based on the narration timestamps. These clips cover a wide range of daily activities and environments, including construction sites, cooking, arts and crafts, shopping, farming, and many others.
Since the timestamp is only an approximate point for where an action occurs, we 
sample the clip from 0.5 s before to 1 s after the timestamp (1.5 s duration) so that the clip is likely long enough to capture the action sound, if there is any, without introducing visuals that stray from the narrated action. 
See ablations on the duration in Supp.
We %sample from the Ego4D training videos and construct a 
sample a training set of 250K clips from 1,876 hours of video. From their narrations, we find there are 6,114 unique nouns (objects) and 2,819 unique verbs (actions). 

\vspace{-0.15in}
\paragraph{Ground truth annotations for evaluation.}

Today's egocentric video datasets lack annotations for sounding actions.  
Thus, to determine how well our model learns long-tail sounding actions and facilitate future research, 
we collect a large ground truth evaluation set \CAcr{for Ego4D} using professional annotators trained for the task.
It consists of 33K clips manually labeled as to whether or not the camera wearer's action sounds, i.e., indicating whether the action described in the narration is both visible and audible in the clip.

To ensure annotation quality, in addition to providing concrete examples and annotation guidelines (see Supp.) and iterating with quality control feedback to the professional annotators,  %to reduce uncertainty, 
we assign three annotators per clip and take the majority vote as the correct answer. 
We split the 33K obtained annotations into 3K for validation and 30K for test. 
We stress that this is an eval set only; our training data (above) has no manual labels about sound, only free-form language narrations.

\begin{table}[t]
\setlength{\tabcolsep}{2pt}
    \centering
    \scalebox{0.95}{
    \begin{tabular}{cccccccccc}
    \toprule
                    Wash & Close & Cut & Drop & Stir & Wipe & Rub & Touch & Lift & Hold \\
    \midrule
                0.90 & 0.82 & 0.77 & 0.64 & 0.64 & 0.53 & 0.39 & 0.27 & 0.19 & 0.09 \\
    \bottomrule
    \end{tabular}}
    \vspace{-0.1in}
    \caption{Example verb groups and %their sounding percentages.
    how frequently they sound}
        \label{tab:action_breakdown}
    \vspace{-0.15in}
    % \vspace{-0.15in}
\end{table}

\vspace{-0.15in}
\paragraph{Action \CAcr{type} analysis.} In total, among the 33,000 resulting ground truth clips, 17,693 are positive and 15,307 are negative.  \KGCR{The fact that only half of this in-the-wild clip distribution consists of sounding actions underscores the need for models that can tell the difference between audio that \emph{co-occurs} with human action and \emph{actions that sound}.} To gain insight into the annotations, we group them by semantic similarity and analyze them at a group level.  While narrations provide semantic descriptions of actions, using them for grouping would be too noisy since the same action could be described in different ways. To reduce the influence of narration variance, we utilize the taxonomy defined in Ego4D \KGCR{(for analysis only, not training)}. For example, ``check", ``examine", and ``inspect" should belong to the same group (taxon). We first group these clips by verb alone, i.e., extracting verbs from narrations and then applying the taxonomy, which results in 106 unique groups. We then compute the percentage of clips in each group that make sounds. \cref{tab:action_breakdown} shows 10 examples.
We see that actions involving more significant human motions (wash, close, cut) are more often sounding, whereas more subtle movements (lift, hold) are often not.  Importantly, there is not a one-to-one mapping between an action verb and its sounding label---how actions sound is scenario-dependent and hence must be mined from the data.

\begin{figure}[t]
    \centering
    \includegraphics[width=\linewidth]{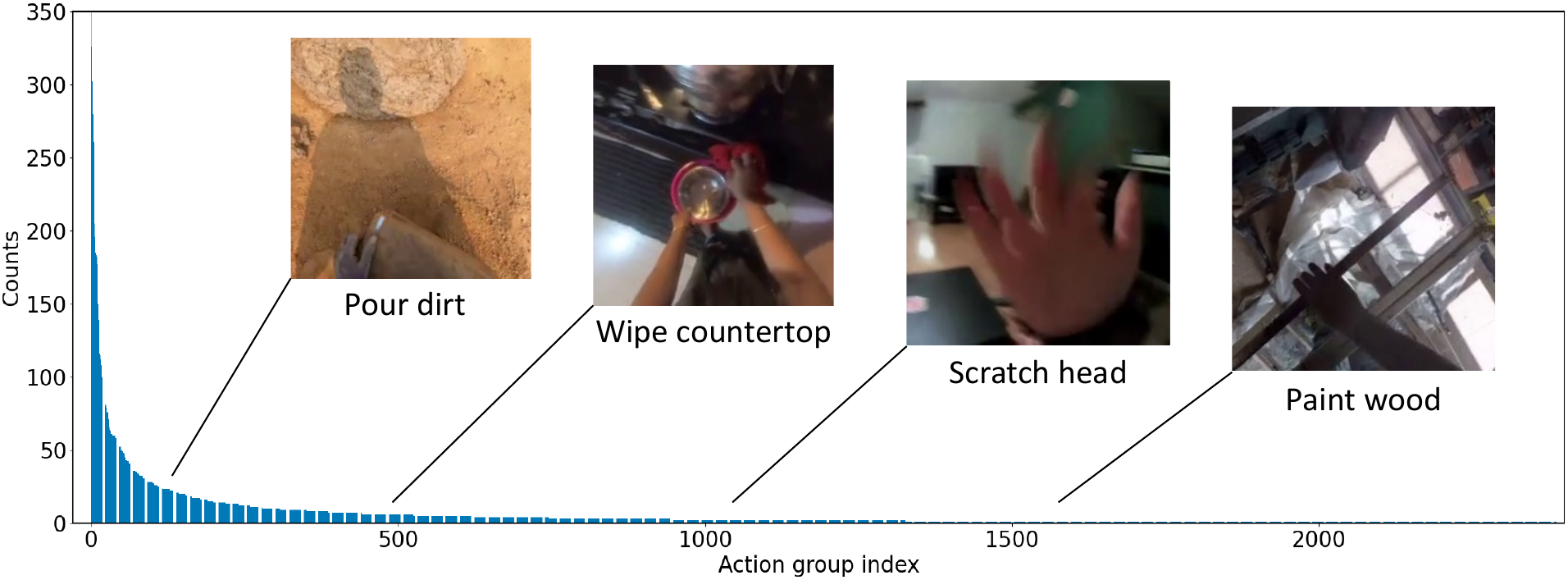}
    \vspace{-0.25in}
    \caption{Long-tail distribution of sounding actions. }
    \vspace{-0.2in}
    \label{fig:long_tail}
\end{figure}

While grouping by verbs provides some insights, how actions make sounds also depends on the object that they interact with, e.g., cutting a carrot sounds different from cutting bread. To this end, we further group the 17K sounding clips by both verbs and nouns, which results in 2,388 unique action groups. We plot the long-tail distribution of them in \cref{fig:long_tail} and show examples sampled from this distribution. This plot shows the diverse and long-tail nature of \CAcr{sounding} actions and our test set annotations, which is not present in existing action datasets~\cite{EPICSOUNDS2023,owens2016vis,Gandhi2020swoosh,clarke2021diffimpact,ucf,caba2015activitynet}. 

%% file: sections/experiments.tex
\section{Experiments} \label{sec:experiments}
\vspace{-0.05in}

We compare our model with several baselines and ablations on three tasks: sounding action discovery \CAcr{(on Ego4D)}, sounding action retrieval (on Ego4D), and audio event classification (on EPIC-Sounds).
We show our model outperforms an array of existing learning methods.

\vspace{-0.15in}
\paragraph{SotA Baselines.} We consider two baselines that only use a contrastive loss for two modalities: CLAP~\cite{clap} for audio-language and CM-ACC~\cite{cm_acc} for audio-video. For more than two modalities, we consider two more baselines: CMC~\cite{cmc} uses contrastive objectives between all pairs of viewpoints (modalities in our case), representing the joint training paradigm; 
ImageBind~\cite{imagebind} learns the joint embedding by first performing vision-language pretraining and then freezing the vision encoder and training the vision-audio modality pair. This represents strategies that align modalities sequentially. For a fair comparison, we equip all baselines with the same encoder and the same initialization as ours (see \cref{sec:implementation}) while keeping their \KGCR{original} losses. 

\begin{table}[t]
\setlength{\tabcolsep}{4pt}
    \centering
    \begin{tabular}{cccccccc}
            \toprule
        & & & & \multicolumn{2}{c}{$AV$} & \multicolumn{2}{c}{$AL$}\\
        & \includegraphics[width=0.03\linewidth]{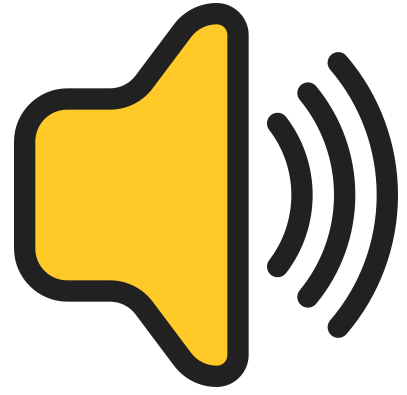} & \includegraphics[width=0.03\linewidth]{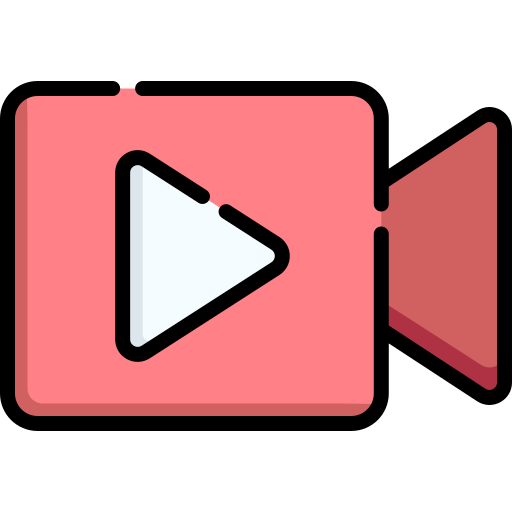} & \includegraphics[width=0.03\linewidth]{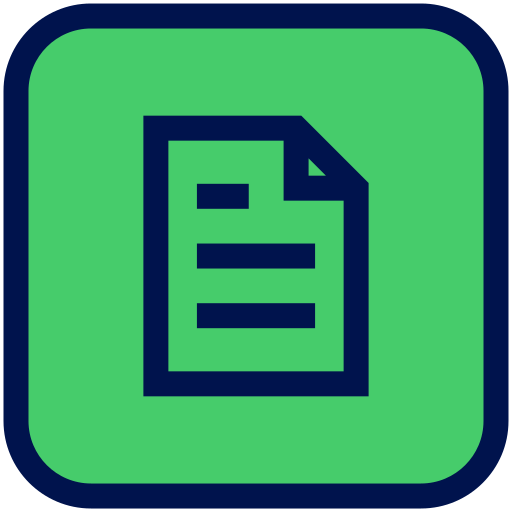}  &  ROC  & PR  & ROC & PR  \\
        \midrule
       Random               & \xmark & \xmark& \xmark & 0.500 & 0.559 & 0.500 & 0.559\\
       CLAP~\cite{clap}     & \cmark & \xmark& \cmark & - & - & 0.637 & 0.695 \\
       CM-ACC~\cite{cm_acc} & \cmark & \cmark& \xmark & 0.540 & 0.590 & - & -  \\
       CMC~\cite{cmc}       & \cmark & \cmark& \cmark & 0.550 & 0.601 & 0.635 & 0.693   \\
       ImageBind~\cite{imagebind} & \cmark & \cmark& \cmark & 0.554 & 0.605 & 0.642 & 0.685  \\
        \midrule
       w/o $\mathcal{L}_{\text{consensus}}$ & \cmark & \cmark& \cmark & 0.563 & 0.615 & 0.635 & 0.694 \\
      w/o $\mathcal{L}_{\text{contrastive}}$ & \cmark & \cmark& \cmark & 0.436 & 0.493 & 0.584 & 0.620 \\
       w/o align-stage     & \cmark & \cmark& \cmark & 0.448 & 0.507 & 0.464 & 0.521 \\
       MC3                 & \cmark & \cmark& \cmark &\tb 0.598 &\tb 0.666 &\tb 0.658 &\tb 0.715\\

        \bottomrule
    \end{tabular}
     \vspace{-0.1in}
    \caption{Sounding action discovery. Area-under-curve (AUC) values are reported for both ROC and precision-recall (PR) curves, \KGCR{for audio-vision (AV) and audio-language (AL)}. Both are the higher the better. We train our model five times with different seeds; the standard deviation is always within $0.01$. 
    }
    \label{tab:discovery}
\vspace{-0.15in}
\end{table}

\begin{figure}[t]
    \centering
    \begin{minipage}{.49\columnwidth}
        \centering
        \includegraphics[width=0.95\linewidth]{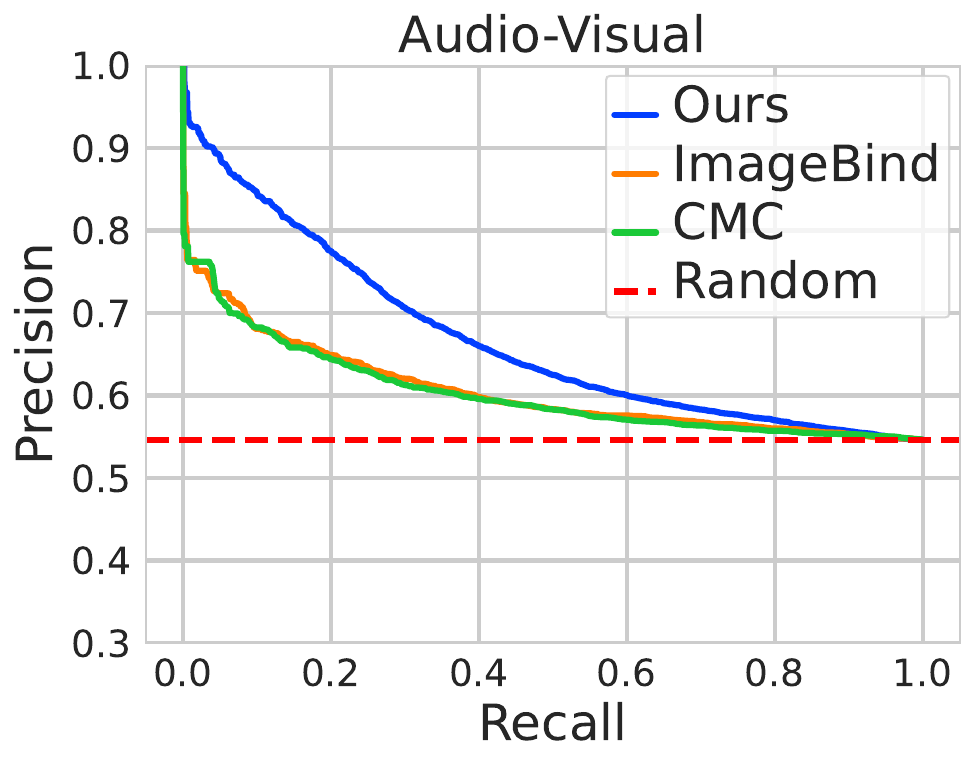} 
        % \caption{Precision-recall curves.}\label{fig:pr_curves}
    \end{minipage}%
    \hfill
    \begin{minipage}{.49\columnwidth}
        \centering
        \includegraphics[width=\linewidth]{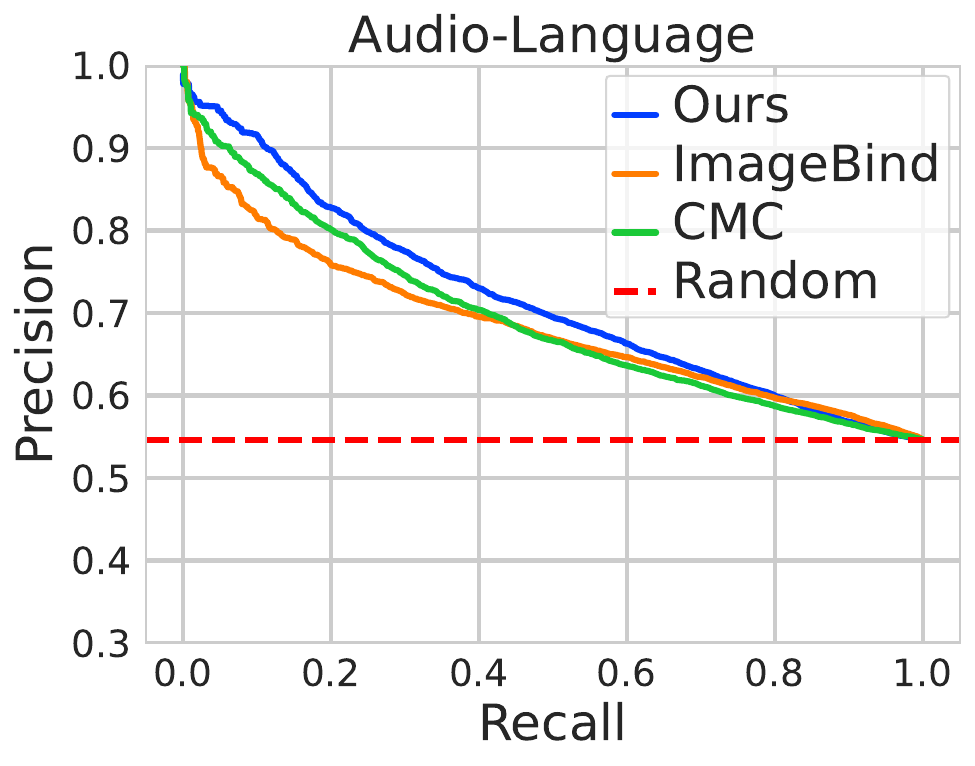}
    \end{minipage}
\vspace{-0.1in}
\caption{\KGCR{Sounding action discovery accuracy} %Precision-recall curves}
}\label{fig:pr_curves}
\vspace{-0.2in}
\end{figure}

\subsection{Sounding Action Discovery}\label{sec:sounding_action_discovery}

Human interactions with objects in our daily lives are complex and subtle. Due to many incidental background sounds, recognizing whether actions make sound is not trivial but can be useful for %automatically mining data from in-the-wild videos for 
applications like multimodal video generation, e.g., verifying the generated action video and audio match.
Towards this goal, we answer the question ``what actions sound?" by performing sounding action discovery. In this experiment, we take the per-modality encoders learned on the narrated 250K Ego4D clips and apply them to the 30K test clips.  Given a test clip,
we feed the video and audio through their corresponding modality encoders, and compute the cosine similarity between the output embeddings. That score indicates how likely it is that the action in the video sounds. 
For completeness, instead of defining a hard threshold for positives, we plot the ROC and precision-recall (PR) curves by varying the positive threshold, and calculate the area-under-curve (AUC) values for them\KGCR{---common metrics for classification~\cite{pr_roc,tharwat2021classification}
 that are invariant to the absolute score values.} For both metrics, higher values are better, indicating the model learns meaningful embeddings of sounding actions.
Similarly, we can also evaluate discovery for audio-language, if narrations are available.

\vspace{-0.15in}
\paragraph{Results.}
Table~\ref{tab:discovery} shows the results for sounding action discovery. 
We first look at discovery with audio-visual modalities alone at test time (``$AV$" columns). CM-ACC~\cite{cm_acc} discovers sounding actions much better than random chance, showing that audio-visual contrastive learning captures both visual action embeddings and action sound embeddings.
CMC~\cite{cmc} and ImageBind~\cite{imagebind} do better---benefiting (like us) from the language modality at training time. However, neither the joint nor sequential training paradigm exploits modality agreement, resulting in weak cross-modal constraints, and thus only marginal performance improvement.
In comparison, our model MC3 explicitly models the modality consensus and improves the discovery result substantially by learning embeddings most relevant to sounding actions.

We also report the discovery result from using audio-language modalities (``$AL$" columns). Since narrations provide action specifications, the discovery performance is better than $AV$, e.g., CLAP~\cite{clap} vs CM-ACC~\cite{cm_acc}. While CMC's~\cite{cmc} and ImageBind's~\cite{imagebind} joint training results are not much better than CLAP~\cite{clap}, our model improves the ``$AL$" discovery by leveraging the video modality and imposing the trimodal consensus constraint.

Fig.~\ref{fig:pr_curves} plots the precision-recall curves.
For the audio-visual curve, our model always has higher precision compared to baselines, especially when recall is low. This is strong evidence of our model learning features of sounding actions, %while 
whereas baselines %capture features for general 
are limited to capturing general
audio-visual correspondence---whether action-based or not. We observe a similar trend for audio-language discovery.

\vspace{-0.2in}
\paragraph{Ablations.} To study the importance of each loss and the two-stage training, we first ablate the consensus loss in the second stage (``w/o $\mathcal{L}_{\text{consensus}}$"  in Table~\ref{tab:discovery}), which trains the model contrastively for both stages. The model performance drops significantly, showing that exploring the modality consensus is key to learning how actions sound.
We then ablate the contrastive loss in the second stage (``w/o $\mathcal{L}_{\text{contrastive}}$"), which harms performance even more. 
This suggests that $\mathcal{L}_{\text{consensus}}$ functions like a regularization term that forces the $\mathcal{L}_{\text{contrastive}}$ to learn sounding action embeddings.
% This shows although the second stage is meant for correcting the mistakes and refining features, jointly optimizing the contrastive loss is still important. 
Lastly, we ablate the two-stage training strategy by removing the align stage (``w/o align-stage"), which optimizes $\mathcal{L}_{\text{MC3}}$ directly; this model fails badly. Aligning embeddings first is critical to making MC3's training stable.
% \KGnoteCR{check if we should elaborate here - does this account for Q3 to MLd8 reviewer?}\CAnote{I think it's pretty self-explanatory, i.e., it's important to perform the first stage then the second stage. the opposite won't work}

\begin{figure}[t]
    \centering
    \includegraphics[width=0.8\linewidth]{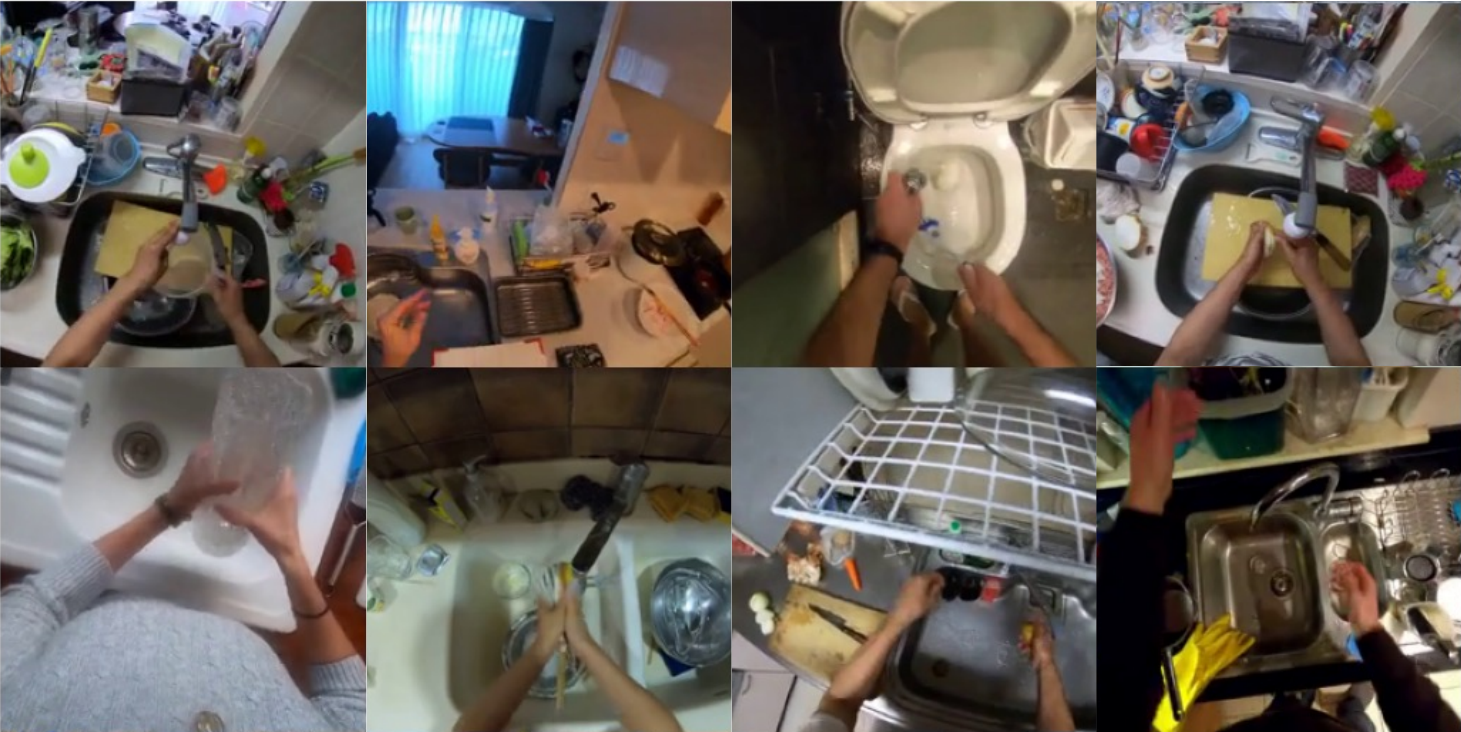}
    \vspace{-0.1in}
    \caption{\KGCR{Example visual embedding cluster from our model}} %Our model's clustered visual embeddings}
    \label{fig:clustering}
    \vspace{-0.2in}
\end{figure}

\vspace{-0.2in}
\paragraph{Clustering.} To visualize the learned embeddings, we group video embeddings in the test set with agglomerative clustering %in the test set 
into 20 clusters.  \cref{fig:clustering} shows the top 8 examples of one cluster (more in Supp.). This cluster clearly captures the sound of water running, \KGCR{despite varying types of camera-wearer movement}. Not only does it group videos with similar actions that make this sound, but also it shows the learned embeddings are agnostic of the background (the bathroom example). %, \CAcr{and unbiased by the head/hand movement since the cluster has varying degrees of movement.}

\begin{table}[t]
    \centering
    \setlength{\tabcolsep}{1pt}
    \begin{tabular}{crrrrrrrrrrrr}
            \toprule
                        & \multicolumn{2}{c}{$V$$\rightarrow$$A$}& \multicolumn{2}{c}{$A$$\rightarrow$$V$}& \multicolumn{2}{c}{$L$$\rightarrow$$A$} & \multicolumn{2}{c}{$A$$\rightarrow$$L$}  \\                        
                        & @5 & @10 & @5 & @10 & @5 & @10 & @5 & @10  \\
        % Recall @         & 1  & 5 & 5 & 1 & 5 & 5 & 1  & 5  & 5 & 1  & 5 & 5\\
        \midrule
       Random            & 0.1 & 0.1 & 0.1 & 0.1 & 0.1 & 0.1 & 0.1 & 0.1\\
       CLAP~\cite{clap}  & - & - & - & - & \tb 49.8 & 87.6 & 34.0 & 67.1\\
       CM-ACC~\cite{cm_acc} & 34.6 & 63.5 & 30.9 & 57.7 & - & - & - & -\\
       CMC~\cite{cmc}    & 36.5 & 67.9 & 33.8 & 63.7 & 44.1 & 81.8 & 32.8 & 64.3\\
       ImageBind~\cite{imagebind} & 32.8 & 61.5 & 29.7 & 57.9 & 42.6 & 76.5 & 30.6 & 60.5\\
        \midrule
       w/o $\mathcal{L}_{\text{consensus}}$ & 33.9 & 63.0 & 30.0 & 56.1 & 45.0 & 84.7 & 32.9 & 65.8\\
        w/o $\mathcal{L}_{\text{contrastive}}$ & 3.3 & 3.7 & 6.4 & 12.5 & 3.1 & 4.7 & 3.3 & 8.0  \\
        w/o align-stage  & 10.0 & 19.4 & 5.9 & 11.8 & 11.6 & 20.9 & 6.5 & 12.6\\
       MC3               &\tb 38.4 &\tb 72.8 &\tb 34.4 &\tb 66.3 & 46.2 &\tb 88.5 &\tb 37.5 &\tb 73.8\\
        \bottomrule
    \end{tabular}
    \vspace{-0.1in}
    \caption{Sounding action retrieval. We report \emph{Recall @5 
 and @10} for different query-retrieval modalities. See R@1 results in Supp.  
    }
    \label{tab:retrieval}
    \vspace{-0.2in}
\end{table}

\begin{figure*}[th!]
    \centering
    \includegraphics[width=\textwidth]{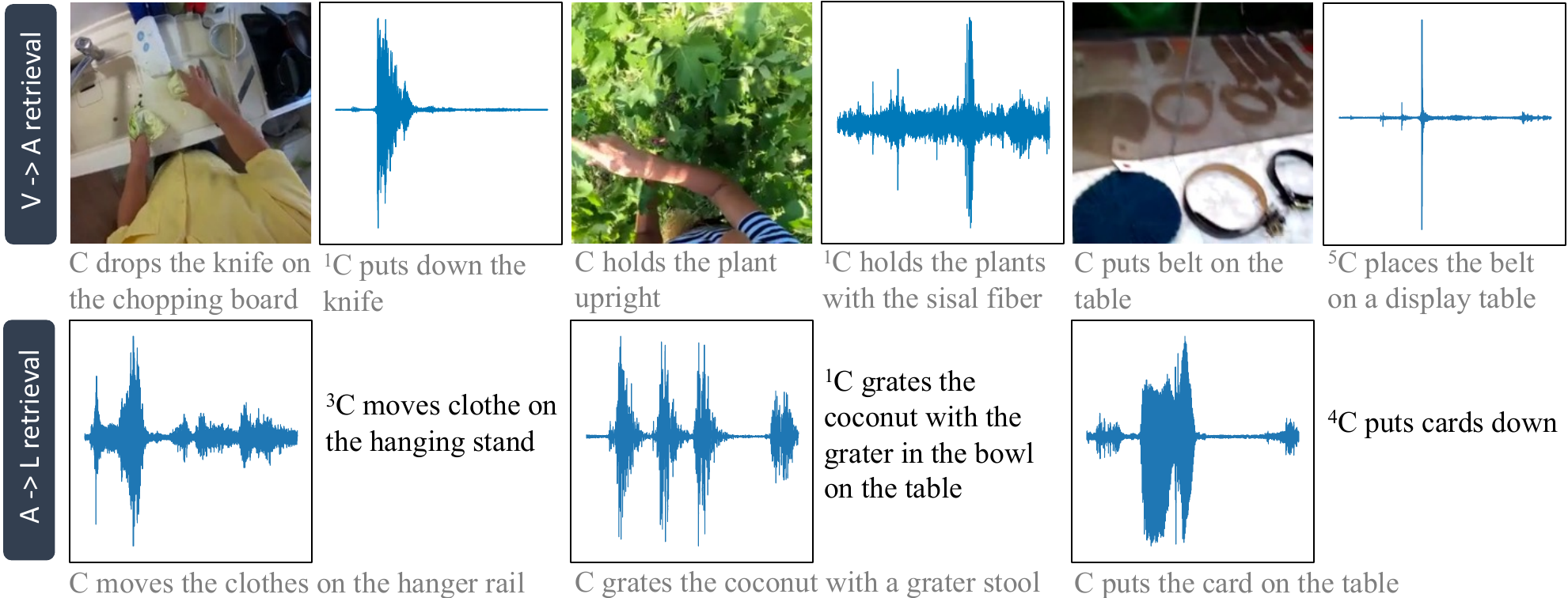}
     \vspace{-0.2in}
    \caption{Qualitative examples for retrieval. The first row is video-to-audio retrieval, motivated by adding audio effects for silent videos. The second row is audio-to-text retrieval, motivated by audio captioning applications. For each row, we show three correct retrieval examples along with their text (gray indicates the text is not observed by the model). For the retrieved item, we show the 
    \KGCR{ground truth} 
    rank as the superscript. All examples are long-tail sounding actions, showing how % that
    our model learns to capture the features of how actions sound.
    }

  \label{fig:qual_retrieval}
  \vspace{-0.1in}
\end{figure*}

% \vspace{-0.1in}
\subsection{Sounding Action Retrieval}\label{sec:sounding_action_retrieval}
Retrieving a different modality given a video, audio, or description is another useful application, such as adding sound effects to silent videos or retrieving captions for action sounds.
To %demonstrate the utility on this application, 
explore this setting, we answer the question ``how do different actions sound?" by evaluating the cross-modal retrieval performance of long-tail sounding actions from the same category. Different from the binary classification task above, here we %look at
aim to retrieve other %retrieving another
examples of the same action. 

To do this, we utilize the action groups constructed in \cref{sec:data} based on verbs and nouns, and only keep groups that have more than two instances of sounding actions
(such that there will be at least one true positive to retrieve for each query). We then divide each action group equally into a query pool of 7,559 examples and a retrieval pool of 7,032 examples.
Given a query modality $M_i$ of instance $A$, we compute its distance to other modalities $M_j$ of all instances in the retrieval pool. A retrieval is correct if the retrieved instance $B$ and $A$ belong to the same action group.

\vspace{-0.2in}
\paragraph{Results.} Table~\ref{tab:retrieval} shows the results 
for four different query-retrieval modality settings. For audio-visual retrieval, we observe that all models can retrieve video with audio (or audio with video) for similar actions with much higher recall than random chance. Our model strongly outperforms the baselines and ablations, benefiting from modeling the modality consensus explicitly. We also observe that retrieving audio with video is easier than the opposite, likely because audio can be vague sometimes, e.g., a collision sound might occur due to various actions while seeing a cutting action indicates the likely sound.
For audio-language retrieval, our model similarly outperforms the baselines by large margins.

% \vspace{-0.2in}
% \ccCR{\paragraph{Different anchor choices.} We also experiment with using video or language as the anchor modality, and find that using audio leads to the best result. See details in Supp.}

\vspace{-0.2in}
\paragraph{Qualitative examples.}  In Fig~\ref{fig:qual_retrieval}, we show examples for video-to-audio and audio-to-language retrieval. \KGCR{Even though} %While
these actions are subtle, our model retrieves audio or captions that are very relevant. Listen to examples in Supp.~video.

% \KGnoteCR{we should address the mLd8 Q1 point somewhere (about body motions/hand movements) from our rebuttal.}\CA{added to supp sec 8.8 where we analyze the visual embeddings}

% \vspace{-0.05in}
\subsection{Audio Classification on EPIC-Sounds}
% \vspace{-0.05in}

%To validate our approach further, we also evaluate the 
Finally, we evaluate our
learned representation on a standard audio benchmark. 
To assess the impact of our model's action sounds representation, we consider EPIC-Sounds~\cite{EPICSOUNDS2023}, a challenging audio classification benchmark for sounds in  kitchen environments.  To our knowledge, EPIC-Sounds represents the only large-scale benchmark for audio in egocentric video.  Note, this classification task is different from the sounding action discovery task in \cref{sec:sounding_action_discovery} in that here the model only takes an audio clip as input.

We consider both linear-probe and fine-tuning settings. In the linear-probe setup, we freeze the model weights and only train the last classification layer, which evaluates the quality of the pre-trained representations. In the fine-tuning setup, we fine-tune both the encoder and the last layer.

Table~\ref{tab:fine_tune} shows the results.
We compare with two SotA methods reported in EPIC-Sounds: SSAST~\cite{ssast} and ASF~\cite{asf}. SSAST is pretrained on LibriSpeech~\cite{librispeech} 
% \ccCR{with self-supervised learning} 
and shares the \emph{same network architecture} as ours, while ASF is trained on VGG-Sound with supervised learning. 
With linear-probe, our model strongly outperforms SSAST~\cite{ssast}, which, 
like us, is also pretrained in a self-supervised fashion with no audio labels. 
ASF~\cite{asf} does better than both, likely due to its advantage of supervised audio classification pretraining.
When fine-tuning,
our model outperforms both prior methods in all but one metric %by more than 2 percent 
\cc{when following the same fine-tuning and evaluation protocol}. This shows our MC3 audio encoder---trained for sounding action discovery---learns generalizable action sound embeddings, improving the state of the art. The margins are naturally smaller in the fine-tuning regime, as is typical, since all models have time to adapt to the new domain.

\begin{table}[!t]
\setlength{\tabcolsep}{4pt}
  \centering
    % \caption{Fine-tuning}\label{tab:fine_tune}
    \begin{tabular}{ccrrrrrr}
        \toprule
                       &  &  Top-1 & Top-5  & mCA & MAP & mAUC \\
    \midrule
   Random             & -  & 7.71 & 30.95 & 2.29 & 0.023 & 0.500 \\
   \rowcolor{gray!20}
   ASF~\cite{asf}*     & L  & 45.53 & 79.33 & 13.48 & 0.172 & 0.789 \\
    SSAST~\cite{ssast} & L & 28.74 & 64.84 & 7.14 & 0.079 & 0.755 \\
   MC3     & L  & 42.44 & 78.76 & 12.79 & 0.153 & 0.818 \\
   \midrule
   \rowcolor{gray!20}
   ASF~\cite{asf}*     & F  & 53.75 & 84.54 & 20.11 & \textbf{0.254} & 0.873 \\
   SSAST~\cite{ssast} & F    & 53.47 & 84.56 & 20.22 & 0.235 & 0.879 \\
    % \midrule
   MC3            & F & \textbf{55.97} & \textbf{85.86} & \textbf{21.65} & 0.242 & \textbf{0.885} \\
    \bottomrule
\end{tabular}
\vspace{-0.05in}
    \caption{Results of classification on EPIC-Sounds. L: Linear-Probe; F: Fine-tuning. * denotes pretraining with supervised audio classification while the rest are pretrained in a self-supervised fashion.  
    }
    \label{tab:fine_tune}
    \vspace{-0.15in}
\end{table}

%% file: sections/conclusion.tex
\vspace{-0.05in}
\section{Conclusion}
\vspace{-0.05in}

We explored %the problem of 
learning how first-person actions sound from in-the-wild, narrated egocentric videos---without audio labels.  
Training with 250K clips from Ego4D, we show the promise of our novel multimodal consensus framework for accurately aligning representations to capture the long-tail of sounding actions in novel (unnarrated) videos, with clear impact on sounding action discovery, retrieval, and pre-training for audio classification.
In the future, we plan to explore multimodal consensus from asynchronous multimodal streams.

\noindent\small{\KGCR{\noindent \textbf{Acknowledgements:} UT Austin is supported in part by the IFML NSF AI Institute. KG is paid as a research scientist by Meta.}}

%% file: sections/supp.tex
\setcounter{section}{7}
\section{Supplementary}

In this supplementary material, we provide additional details about:
\begin{enumerate}
    \item Supplementary video for qualitative examples (referenced in Sec.~6).
    \item Annotation guidelines and interface (referenced in Sec.~5).
    % \item Retrieval evaluation details (referenced in Sec.~5).
    \item Additional implementation details (referenced in Sec.~5).
    \item Ablations on anchor modality (referenced in Sec.~6).
    \item Ablations on hyperparameter  (referenced in Sec.~4).
    \item Ablations on the time window length  (referenced in Sec.~5).
    \item Recall @1 for sounding action retrieval  (referenced in Sec.~6).
    \item More clusters of visual embeddings (referenced in Sec.~6).
    
\end{enumerate}

\subsection{Supplementary Video}
In this video, we include examples of Ego4D clips, qualitative examples of sounding action discovery, and examples of sounding action retrieval. Wear headphones to hear the sound.

\subsection{Annotation Guidelines and Interface}
\begin{figure*}[t]
    \centering
    \includegraphics[width=\linewidth]{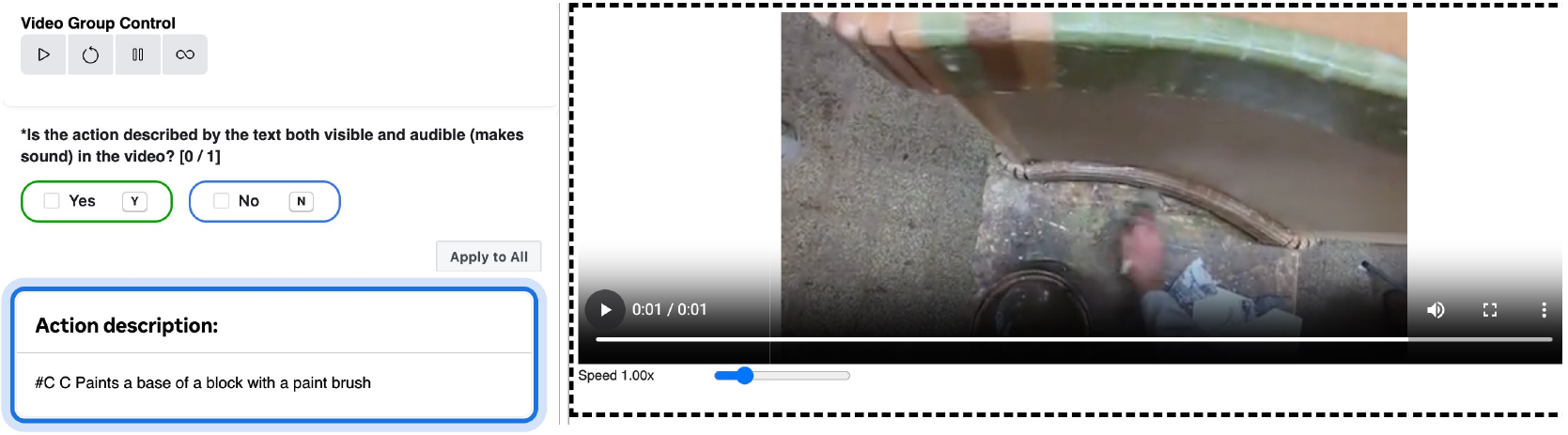}
    \vspace{-0.2in}
    \caption{Annotation interface.}
    \vspace{-0.1in}
    \label{fig:halo_interface}
\end{figure*}

For annotators, we first tried MTurk, which we found too noisy. To get high-quality annotations, we then hired 8 professional annotators to work on the annotation task. Each instructor received annotation training and read the annotation guidelines before annotating. They are instructed to classify whether the foreground action described by the narration is both visible and audible in the clip. We also provided them with some positive examples and negative examples to start with. Fig.~\ref{fig:halo_interface} shows the annotation interface.

% \subsection{Retrieval Evaluation Details}
% In the sounding action retrieval experiment, we broadly define action classes based on verbs and verb taxonomy. The 11 actions and their associated labels are: ``wash", ``close", ``open", ``cut(trim,slice,chop)", ``throw(toss,dump,dispose)", ``put(place,leave,drop)", ``press", ``mix(stir)", ``scoop", ``arrange(straighten,sort,distribute,align)", ``wipe". The full list of verbs under these labels can be looked up in the ``narration\_verb\_taxonomy.csv" supplementary file.

% We also consider a more fine-grained evaluation ``grouping by verb and noun", where a retrieval is correct if and only if both verbs and nouns belong to the same verb/noun groups. And the taxonomy for nouns can be found in the ``narration\_noun\_taxonomy.csv" supplementary file.

\subsection{Additional Implementation Details}

% \paragraph{Baseline implementation.} To investigate whether our model does more than weighting bad samples, we've equipped the weighting strategy~\cite{morgado2021_robust_xid} for CMC~\cite{cmc}, where we estimate the weight for samples based on their similarity score. We followed Eq.~6 in ~\cite{morgado2021_robust_xid} and used the best hyperparameters discovered in their paper for shaping the Gaussian distribution.

Following the setting of previous work~\cite{kevin2022egovlp}, we initialize the video encoder with ViT~\cite{vit} pretrained on ImageNet~\cite{deng2009imagenet} that has a latent dimension of $768$. We use the ``distilbert-base-uncased" transformer from Huggingface as our text encoder, which has a latent dimension of 256. For audio encoder, we use AST~\cite{gong21b_interspeech} that has been initialized with ViT~\cite{vit} pretrained on ImageNet~\cite{deng2009imagenet}. For the joint embedding space, we project features of audio, video and text into a latent space with dimension $256$. During training, we resize the video to $224\times224$ and use $4$ frames per clip. For audio, we use a sample rate of $16000$. We extract fbank features from the audio waveform with $128$ Mel frequency bins, $10$ ms frame shift and hanning windows.

\subsection{Ablations on Anchor Modality}
To study the importance of the choice of the anchor modality, we experiment with using video or language as the anchor and report the retrieval performance in Table~\ref{tab:anchor_abalation}. Using video or language as the anchor modality has a similar but slightly lower performance compared to anchoring audio, likely because audio is generally more ambiguous and thus benefits more from being used as the anchor modality.

\begin{table}[t]
    \centering
    \setlength{\tabcolsep}{1pt}
    % \begin{tabular}{ccccccccccccc}
    \begin{tabular}{crrrrrrrrrrrr}
    % \begin{tabular}{cS[table-format=2.1]S[table-format=2.1]S[table-format=2.1]S[table-format=2.1]S[table-format=2.1]S[table-format=2.1]S[table-format=2.1]S[table-format=2.1]S[table-format=2.1]S[table-format=2.1]S[table-format=2.1]S[table-format=2.1]}
            \toprule
                        & \multicolumn{2}{c}{$V$$\rightarrow$$A$}& \multicolumn{2}{c}{$A$$\rightarrow$$V$}& \multicolumn{2}{c}{$L$$\rightarrow$$A$} & \multicolumn{2}{c}{$A$$\rightarrow$$L$}  \\                        
        MC3   & @5 & @10 & @5 & @10 & @5 & @10 & @5 & @10  \\
        \midrule
       Audio as anchor  &\tb 38.4 &\tb 72.8 &\tb 34.4 &\tb 66.3 & 46.2 & 88.5 &\tb 37.5 &\tb 73.8\\
      Video as anchor   & 38.1 & 72.4 & 31.9 & 62.5 &\tb 46.6 &\tb 88.7 & 36.3 & 70.7 \\
      Language as anchor   & 37.1 & 70.0 &\tb 34.4 & 66.0 & 45.7 & 84.9 & 29.6 & 61.2\\

        \bottomrule
    \end{tabular}
    \caption{Ablations on the anchor modality.  
 % \KGnote{should gray anchor part go in supp? also label more clearly, like ``MC3 w/V anchor"?}
 % We also evaluate two settings here: VB indicates a correct retrieval happens when the verbs of the query and retrieval are within the same group in the verb taxonomy while VN indicates both the verb and noun have to match. 
    % \KGnote{this table and table 2 are large.  could we reduce the number of columns in this one to make more space for qualitative figures?}
    }
    \label{tab:anchor_abalation}
    \vspace{-0.05in}
\end{table}

\begin{figure*}[t]
    \centering
    \begin{subfigure}[b]{0.33\textwidth}
        \includegraphics[width=\textwidth]{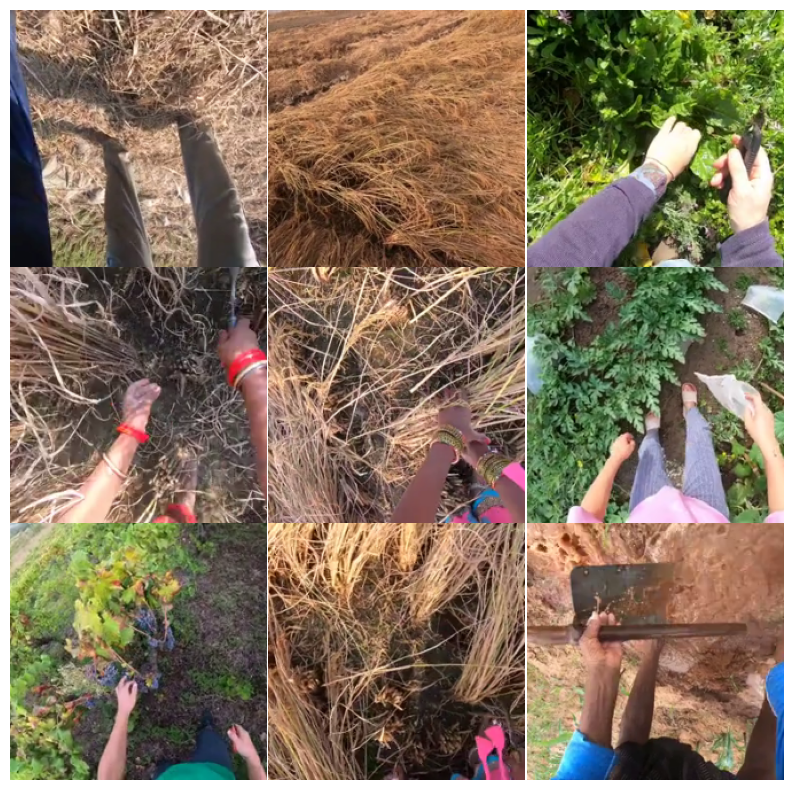}
        \caption{Actions that make the rustle sound.}
        \label{fig:rustle}
    \end{subfigure}
    \hfill
    \begin{subfigure}[b]{0.33\textwidth}
        \includegraphics[width=\textwidth]{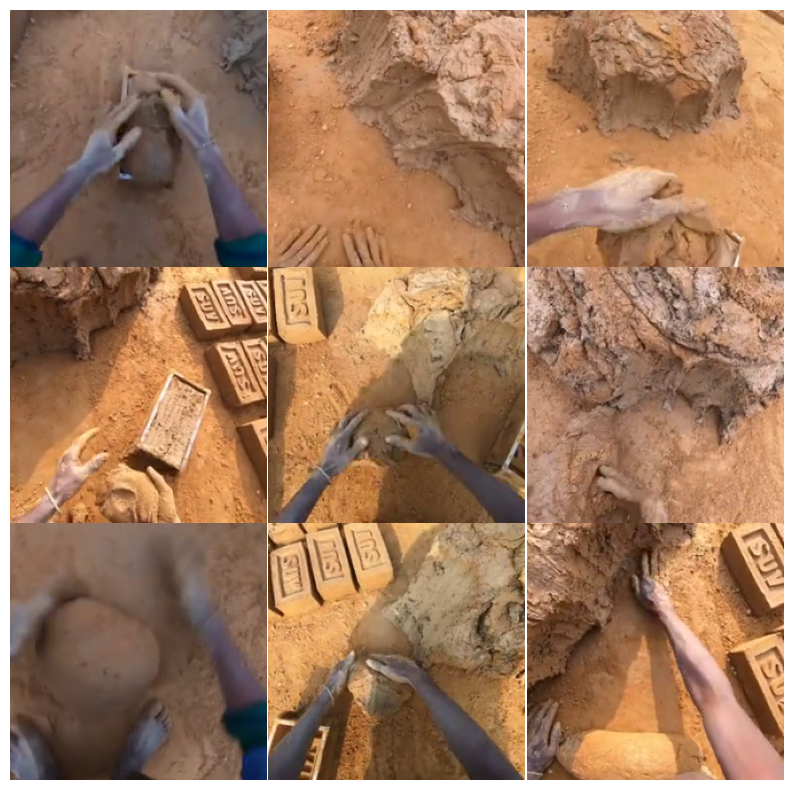}
        \caption{Actions that scoop the mud.}
        \label{fig:scoop}
    \end{subfigure}
    \hfill
        \begin{subfigure}[b]{0.33\textwidth}
        \includegraphics[width=\textwidth]{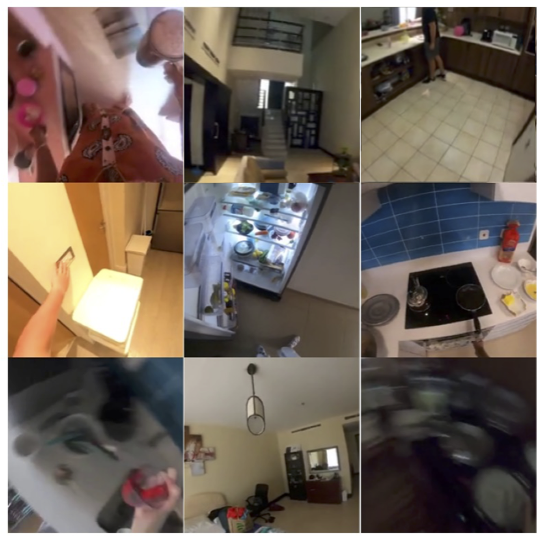}
        \caption{Actions that make footstep sound.}
        \label{fig:walking}
    \end{subfigure}
    % \hfill
    % \begin{subfigure}[b]{0.24\textwidth}
    %     \includegraphics[width=\textwidth]{figures/blurred_conversation.pdf}
    %     \caption{Videos not sound due to actions.}
    %     \label{fig:conversation}
    % \end{subfigure}
    \vspace{-0.1in}
    \caption{More clusters of visual embeddings.}
    \vspace{-0.1in}
    \label{fig:more_clusters}
\end{figure*}

\begin{table}[t]
    \centering
    \setlength{\tabcolsep}{4pt}
    \begin{tabular}{l|rrrrrrrrrrrr}
            \toprule
                        & \multicolumn{2}{c}{$V$$\rightarrow$$A$}& \multicolumn{2}{c}{$A$$\rightarrow$$V$}& \multicolumn{2}{c}{$L$$\rightarrow$$A$} & \multicolumn{2}{c}{$A$$\rightarrow$$L$}  \\                        
        $\alpha_v$   & @5 & @10 & @5 & @10 & @5 & @10 & @5 & @10  \\
        \midrule
       0.1   & 37.5 & 71.5 & 34.0 & 64.7 & 46.2 & 87.9 & 36.2 & 71.5\\
       0.25   & 37.9 & 72.2 & 33.9 & 66.1 &\tb 47.4 & 87.2 & 36.1 & 71.5\\
       0.5   &\tb 38.4 &\tb 72.8 &\tb 34.4 &\tb 66.3 & 46.2 &\tb 88.5 &\tb 37.5 &\tb 73.8\\
       0.75   & 37.2 & 70.7 & 34.1 & 65.8 & 44.2 & 86.4 & 36.9 & 71.4\\
       1.0   & 37.8 & 70.6 & 32.4 & 62.8 & 47.3 &\tb 88.5 & 35.4 & 70.4\\
       2.0   & 27.6 & 52.5 & 16.4 & 35.2 & 43.5 & 82.0 & 13.5 & 24.9\\

        \bottomrule
    \end{tabular}
    \caption{Ablations on the hyperparameter. }
    \vspace{-0.1in}
    \label{tab:anchor_hyperparameter}
    % \vspace{-0.15in}
\end{table}

\subsection{Ablations on Hyperparameters}
To make scores from different modality pairs comparable, we use $\mathcal{K}_i(x) = ((x+1)/2)^{\alpha_i}$ to adjust the distribution. Since we set audio as the anchor modality, we only need to tune the $\alpha_V$ and $\alpha_L$. For tuning, we first set $\alpha_L$ to 1 and then perform a grid search of $\alpha_V$ on the validation data. We report the retrieval performance of all values in \cref{tab:anchor_hyperparameter}.
We chose 0.5 as the weight since it achieves the best performance. 

\begin{table}[t]
    \centering
    \setlength{\tabcolsep}{4pt}
    \begin{tabular}{l|rrrrrrrrrrrr}
            \toprule
                        & \multicolumn{2}{c}{$V$$\rightarrow$$A$}& \multicolumn{2}{c}{$A$$\rightarrow$$V$}& \multicolumn{2}{c}{$L$$\rightarrow$$A$} & \multicolumn{2}{c}{$A$$\rightarrow$$L$}  \\                        
               & @5 & @10 & @5 & @10 & @5 & @10 & @5 & @10  \\
        \midrule
       0.5 s  & 26.0 & 49.0 & 18.0 & 36.2 & 32.0 & 58.5 & 20.3 & 39.9 \\
       1.0 s  & 32.8 & 62.3 & 28.7 & 54.8 & 42.1 & 80.1 & 32.1 & 62.2 \\
       1.5 s  &\tb 38.4 &\tb 72.8 &\tb 34.4 &\tb 66.3 & \tb 46.2 & \tb 88.5 &\tb 37.5 &\tb 73.8\\
       2.0 s  & 34.3 & 64.2 & 30.8 & 60.7 & 37.2 & 71.0 & 29.8 & 58.8 \\
        \bottomrule
    \end{tabular}
    \caption{Ablations on the time window length.}
    \label{tab:time_abalation}
    \vspace{-0.1in}
\end{table}

\subsection{Ablations on Time Window Length}
Narrations are timestamped and the action sound (if any) happens within a time window of the timestamp. For the duration of the time window, we consider a few values (0.5 s, 1.0 s, 1.5 s, and 2.0 s) and report their retrieval performance in \cref{tab:time_abalation}. Choosing a 1.5 s window leads to the best performance, which is likely because too short time windows can often miss the action sound while long time windows would introduce noise or other action sounds. However, our model is not super sensitive to the choice of the window length since it also performs well with other lengths.

\begin{table}[t]
    \centering
    \setlength{\tabcolsep}{2pt}
    \begin{tabular}{crrrrrrrrrrrr}
            \toprule
                        & \multicolumn{2}{c}{$V$$\rightarrow$$A$}& \multicolumn{2}{c}{$A$$\rightarrow$$V$}& \multicolumn{2}{c}{$L$$\rightarrow$$A$} & \multicolumn{2}{c}{$A$$\rightarrow$$L$}  \\                        
                        & @1 & @5 & @1 & @5 & @1 & @5 & @1 & @5  \\
        \midrule
       Random            & 0.0 & 0.1 & 0.0 & 0.1 & 0.0 & 0.1 & 0.0 & 0.1\\
       CLAP~\cite{clap}  & - & - & - & - & 10.8 &\tb 49.8 & 6.5 & 34.0\\
       CM-ACC~\cite{cm_acc} & 7.7 & 34.6 & 6.7 & 30.9 & - & - & - & -\\
       CMC~\cite{cmc}    & 7.6 & 36.5 &\tb 7.6 & 33.8 & 9.7 & 44.1 & 6.7 & 32.8\\
       ImageBind~\cite{imagebind} & 7.2 & 32.8 & 6.1 & 29.7 & 8.6 & 42.6 & 5.9 & 30.6\\
        \midrule
       % w/o $\mathcal{L}_{\text{consensus}}$ & 33.9 & 63.0 & 30.0 & 56.1 & 45.0 & 84.7 & 32.9 & 65.8\\
       %  w/o $\mathcal{L}_{\text{contrastive}}$ & 3.3 & 3.7 & 6.4 & 12.5 & 3.1 & 4.7 & 3.3 & 8.0  \\
        % \CAr{w/o align-stage } & 10.0 & 19.4 & 5.9 & 11.8 & 11.6 & 20.9 & 6.5 & 12.6\\
       MC3               & \tb7.8 &\tb 38.4 & 7.2 &\tb 34.4 &\tb 11.3 & 46.2 &\tb 7.3 & \tb37.5\\
        \bottomrule
    \end{tabular}
    \caption{Sounding action retrieval. We report \CAr{\emph{Recall @1 
 and @5}} for different query-retrieval modalities.
    }
    \label{tab:retrieval_supp}
    \vspace{-0.1in}
\end{table} 

\subsection{Recall @1 for Sounding Action Retrieval}
Due to the space limit in the main, we report Recall @1 for the retrieval experiment in \cref{tab:retrieval_supp}. While the performance gap is small as expected, our model still outperforms baselines on most of the metrics.

\subsection{More Clusters of Visual Embeddings}
In Sec.~6 of the main, we showed one cluster of visual embeddings. Here we show three more clusters from the same clustering result in \cref{fig:more_clusters}. \cref{fig:rustle} clusters visual actions that make rustle sounds when interacting with grass/branches, even though some examples have very different backgrounds (yellow vs green). This indicates our model learns to cluster visual actions based on how they sound rather than just how they look. \cref{fig:scoop} shows a cluster of visual actions that scoop the mud/dirt. \cref{fig:walking} shows the visual cluster where the walking action produces footsteps. 
% \cref{fig:conversation} shows the visual cluster of social interactions that likely do not make sound due to visible camera wearer's actions since their speaking activity cannot be inferred from visual alone. 
\CAcr{Each cluster has actions with varying degrees of head and hand movement, and our model still captures accurately how actions make sounds despite the movement.}